\documentclass{article} 
\usepackage{iclr2026_conference,times}

\usepackage{amsmath,amsfonts,bm}

\def\eqref#1{equation~\ref{#1}}

\def\1{\bm{1}}

\DeclareMathAlphabet{\mathsfit}{\encodingdefault}{\sfdefault}{m}{sl}
\SetMathAlphabet{\mathsfit}{bold}{\encodingdefault}{\sfdefault}{bx}{n}

\usepackage{hyperref}
\usepackage{url}
\usepackage{booktabs}
\usepackage{multirow}
\usepackage{makecell}
\usepackage{hyperref}
\usepackage{url}
\usepackage{amsmath}
\usepackage{graphicx}
\usepackage{pifont}
\title{OccVLA: Vision-Language-Action Model with Implicit 3D Occupancy Supervision}

\author{Ruixun Liu\textsuperscript{1,2}\footnotemark[1]~, Lingyu Kong\textsuperscript{1,3}\footnotemark[1]~,  Derun Li\textsuperscript{1,4}\thanks{Equal contribution.}, Hang Zhao\textsuperscript{1,5}\thanks{Corresponding author. hangzhao@mail.tsinghua.edu.cn} \\
\textsuperscript{1}Shanghai Qi Zhi Institute, \textsuperscript{2}Xi'an Jiaotong University, \textsuperscript{3}Fudan University\\
\textsuperscript{4}Shanghai Jiao Tong University, 
\textsuperscript{5}Tsinghua University
\\
}

\iclrfinalcopy 
\begin{document}

\maketitle

\begin{abstract}
Multimodal large language models (MLLMs) have shown strong vision–language reasoning abilities but still lack robust 3D spatial understanding, which is critical for autonomous driving. 
This limitation stems from two key challenges: (1) the difficulty of constructing accessible yet effective 3D representations without expensive manual annotations, and (2) the loss of fine-grained spatial details in VLMs due to the absence of large-scale 3D vision–language pretraining.
To address these challenges, we propose OccVLA, a novel framework that integrates 3D occupancy representations into a unified multimodal reasoning process. Unlike prior approaches that rely on explicit 3D inputs, OccVLA treats dense 3D occupancy as both a predictive output and a supervisory signal, enabling the model to learn fine-grained spatial structures directly from 2D visual inputs. 
The occupancy prediction are regarded as implicit reasoning processes and can be skipped during inference without performance degradation, thereby adding no extra computational overhead.
OccVLA achieves state-of-the-art results on the nuScenes benchmark for trajectory planning and demonstrates superior performance on 3D visual question-answering tasks, offering a scalable, interpretable, and fully vision-based solution for autonomous driving.
\end{abstract}

\section{Introduction}
\begin{figure}[t]
  \centering
   \includegraphics[width=1.0\linewidth]{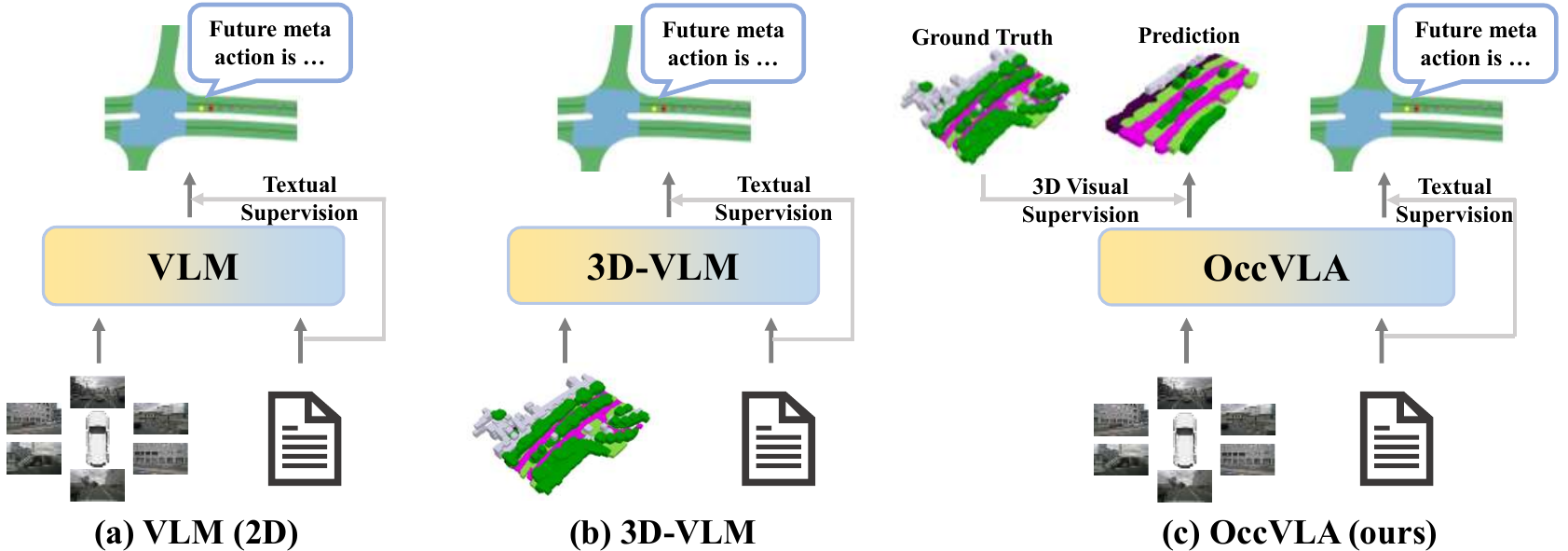}
   \caption{Comparison of autonomous driving VLM architectures.
(a) VLM (2D): Takes only 2D visual inputs and relies solely on textual supervision, lacking explicit 3D spatial grounding.
(b) 3D-VLM: Consumes explicit 3D inputs (e.g., Occupancy,  LiDAR) for reasoning, but the absence of large-scale 3D vision–language pretraining often leads to loss of fine-grained spatial details and limits generalization.
(c) OccVLA (ours): Predicts dense 3D occupancy from 2D images and uses it as both an output and a dense 3D supervisory signal, enhancing fine-grained spatial understanding while preserving rich 2D visual details.}
   \label{intro}
   \vspace{-4mm}
\end{figure}

 Recently, end-to-end autonomous driving \citep{hu2022stp3endtoendvisionbasedautonomous, jiang2023vadvectorizedscenerepresentation, contributors2023_uniadrepo, hu2023_uniad} has witnessed remarkable advances, driven by increasing demands for real-world deployments. Advanced autonomous driving systems  \citep{zhou2025opend, zheng2025drive} now routinely integrate vision language models (VLMs) to deliver compelling reasoning capabilities in complex driving scenarios. 
 Nevertheless, the persistent gap between 2D and 3D perception remains a principal limitation to broader VLM adoption. In autonomous driving, robust 3D perception \citep{qi2017pointnet, lang2019pointpillars, wang2022detr3d} is indispensable for localization and navigation, since its fidelity directly influences the safety of downstream decision-making. 
 Prior work has extensively explored this challenge as shown in Fig.~\ref{intro} (a). 
 In VLM-based perception pipelines \citep{tian2024drivevlm, hwang2024emma}, supervision relies on 3D annotations described in text (e.g., coordinates or bounding boxes), which are inherently weak and sparse. Producing such annotations demands extensive manual labeling, thereby constraining scalability.
 As illustrated in Fig.~\ref{intro} (b), some recent methods \citep{wang2025omnidrive, wei2024occllama, xiong2023neuralmappriorautonomous} attempt to incorporate 3D inputs, but they are limited by the lack of large-scale 3D vision–language pretraining and detailed captions for complex spatial scenes. 
 Such 3D VLMs generally focus on supervising textual outputs while overlooking the rich 3D visual modality, leaving potential for improving spatial understanding in autonomous driving.

Two critical challenges arise in this context: 
(1) establishing an accessible and effective representation of 3D information, and 
(2) developing dense 3D supervision to preserve fine-grained spatial details. 
Recent progress in automated annotation pipelines \citep{NEURIPS2023_cabfaeec, ye2025gs} enables large-scale acquisition of 3D occupancy representations for autonomous driving scenarios. Such representations naturally encode both detailed 3D structural geometry and semantic labels, providing a unified format for aligning spatial and semantic information. 
With advancements in occupancy prediction techniques, transformer-based models \citep{li2023voxformer, huang2023triperspectiveviewvisionbased3d, zhang2023occformer} have demonstrated their feasibility for modeling this representation. 
Inspired by these developments \citep{li2023fbbev, li2023stereoscene}, we propose a VLM augmented with occupancy prediction capabilities, to simultaneously address the representation and supervision challenges.

\begin{figure}[t]
  \centering
   \includegraphics[width=1.0\linewidth]{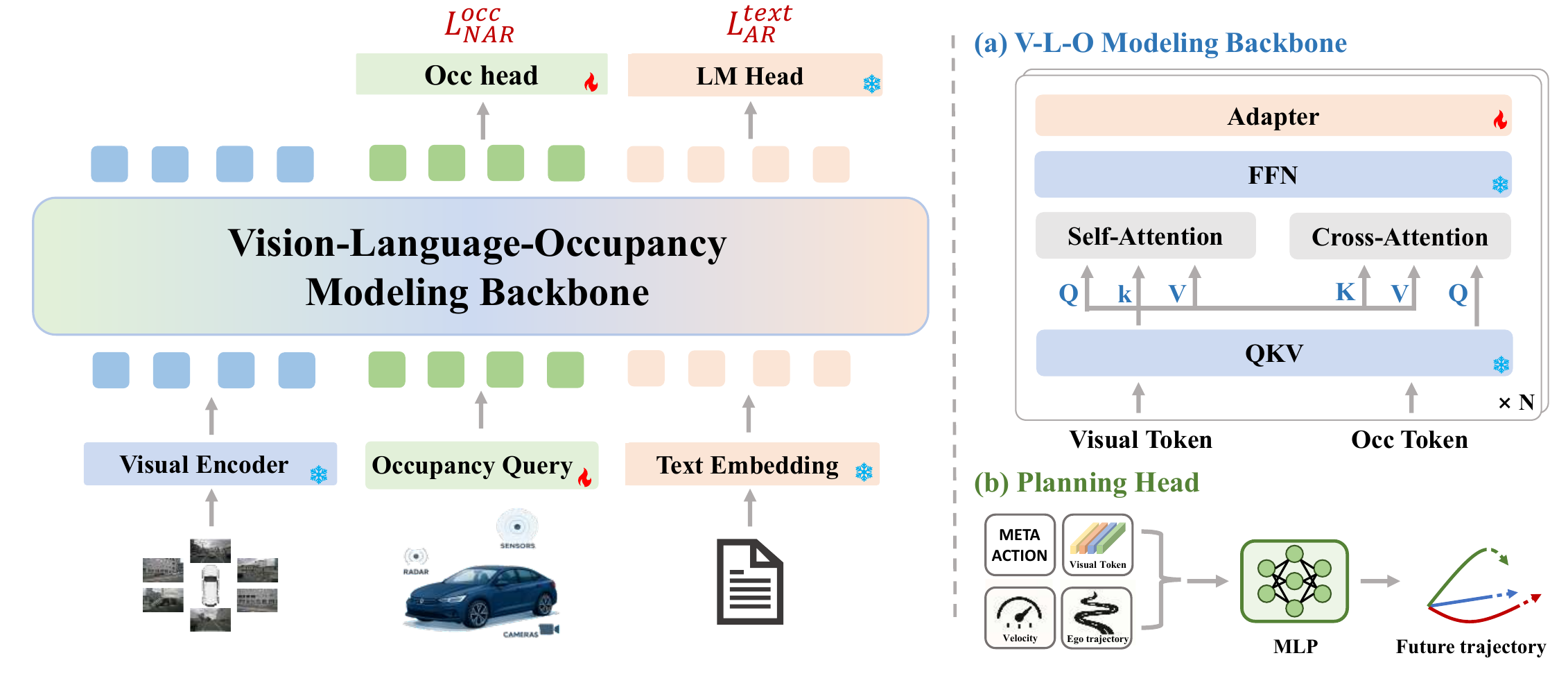}
   \caption{Overview of the proposed OccVLA architecture. The framework unifies dense 3D occupancy (occ) prediction and language modeling within a shared Vision–Language–Occupancy (V-L-O) backbone.
   The model is jointly trained with $L_{NAR}^{occ}$ (a non-autoregressive loss for occupancy prediction) and $L_{AR}^{text}$ (an autoregressive loss for textual outputs). (a) In the V-L-O backbone, occupancy tokens query visual features from visual tokens through cross-attention, while visual tokens are modeled via self-attention. (b) After predicting meta actions through the VLM, a planning head (MLP) generates the future trajectory.}
   \label{flowchart}
   \vspace{-4mm}
\end{figure}

Building on this perspective, we introduce a novel framework, \textbf{Occ}upancy \textbf{V}ision-\textbf{L}anguage-\textbf{A}ction model (OccVLA), which enables execution of occupancy prediction, vision-language reasoning and action generation. 
As illustrated in Fig.~\ref{flowchart}, OccVLA treats occupancy tokens as implicit reasoning processes, using cross-attention to receive visual features from intermediate layers of the VLM.
To address the spatial sparsity of occupancy representations \citep{wei2024occllama}, we first predict occupancy in a compact latent space, after which an occupancy head maps the resulting occupancy tokens back to the high-resolution original occupancy space. 
This 3D scene prediction step enables the VLM to capture fine-grained spatial details more effectively. 
Moreover, compared to raw visual features, supervising on the occupancy representation substantially enhances the 3D representational capacity of the VLM’s visual features. 
Notably, during inference, the occupancy prediction process can remain inactive, introducing no additional computational overhead. 
Finally, a lightweight MLP consumes the meta-actions predicted by the VLM to predict future trajectories, providing a simple yet effective solution for trajectory forecasting.

OccVLA demonstrates superior performance across multiple perception and planning tasks. We validate its 3D understanding capabilities on the nuScenes dataset through various VQA tasks \citep{qian2023nuscenes, inoue2024nuscenes}, such as relative vehicle position localization. The visual input to OccVLA consists of only 2D images, which effectively preserves the inherent generalization capability of VLMs during open-domain dialogue. Notably, OccVLA offers the flexibility to decode the occupancy representation, producing interpretable and quantitatively evaluable outputs, which is particularly advantageous for fully vision-based autonomous driving solutions.

The main contributions of this paper are as below:
\begin{enumerate}
    \item We propose the autonomous driving framework OccVLA, which extends the 3D reasoning capabilities of vision-language models (VLMs) through the occupancy prediction process while effectively preserving visual information from 2D images.
    \item The design of the cross-modal attention allows the model to skip the occupancy prediction process during inference, introducing no additional computational complexity.
    \item OccVLA achieves outstanding performance in both end-to-end trajectory planning and 3D VQA tasks, setting state-of-the-art results on the public benchmark nuScenes.
\end{enumerate}

\section{Related Work}

\subsection{MLLMs in Autonomous Driving}   
Recent studies \citep{sima2023drivelm, wang2023drivemlmaligningmultimodallarge, zhang2025mpdrive} argue that multimodal large language models (MLLMs) can emulate the human thought process during driving. Leveraging the exceptional zero-shot generalization capabilities of vision-language models (VLMs) \citep{tian2024drivevlm, xu2024drivegpt4interpretableendtoendautonomous}, they can effectively handle long-tail scenarios in autonomous driving. However, due to limitations in their pretraining paradigms, VLMs struggle to effectively comprehend the 3D structure of the physical world. DriveVLM \citep{tian2024drivevlm} is the first to propose using VLMs for autonomous driving motion planning, but it relies on high-quality annotated datasets. EMMA \citep{hwang2024emma} employs extensive datasets containing 3D coordinates to enhance the model's 3D grounding capabilities, but this approach requires significant manual annotation efforts. Similarly, OmniDrive \citep{wang2025omnidrive} compresses 3D point clouds into sparse queries and feeds them into large language models (LLMs), which necessitates additional 3D sensors and forces the model to process large-scale 3D inputs. In this work, we propose OccVLA, which leverages auto-annotation occupancy data to provide dense 3D supervision for MLLMs.

\subsection{Occupancy for 3D Perception}
3D occupancy assigns semantic labels to spatial grids, aiming to establish fine-grained representations of 3D scenes. Transformer-based methods \citep{liu2024occtransformer, li2024bevformer}, through spatiotemporal feature fusion, have demonstrated significant advantages in occupancy prediction tasks. Recently, unlike traditional vision-language models (VLMs), several studies have explored the potential of using occupancy as input of LLM to enhance the understanding capabilities of multimodal large language models (MLLMs) in autonomous driving. OccWorld \citep{zheng2024occworld} proposes making predictions on multi-scale occupancy features to learn a world model, while OccLLAMA \citep{wei2024occllama} introduces the use of large language models (LLMs) to predict future 3D occupancy and actions. Similarly, Occ-LLM \citep{xu2025occ} proposes a motion-separating variational autoencoder that disentangles dynamic and static objects in occupancy grids and predicts them separately using LLMs.
Although it is possible to perform joint training of 3D visual inputs and language similar to VLMs, there remains a risk that captions omit critical 3D information. To address these limitations, OccVLA focuses on using occupancy as both the model's output and supervision signal, thereby establishing a novel framework for multimodal learning.

\section{Method}
\subsection{Overview}
In this section, We propose OccVLA, a unified framework for 3D occupancy prediction and future ego-motion planning. The core components of OccVLA include the occupancy prediction (Section 3.2) and an independent planning head (Section 3.3). Additionally, we introduce a three-stage training process (Section 3.4) to better balance the model's performance across different tasks.

We incorporate 3D visual supervision into the typical VLM framework, as illustrated in Fig.~\ref{flowchart}. Before performing next-token prediction, the model first perceives the visual input and produces an occupancy prediction. This unified architecture enables seamless integration of visual and textual information during the perception stage (perceive first, then reason), thereby establishing a solid perceptual foundation for visual understanding, mitigating the information loss caused by text-only supervision, and ultimately enhancing the model’s 3D comprehension capability.



\subsection{Occupancy Prediction}

\textbf{Occupancy Transformer.} To strengthen the 3D perception capability of autonomous driving systems,  we extend the original VLM framework with a dedicated 3D occupancy prediction processing. OccVLA takes a set of learnable occupancy queries as input, which are passed through the same feed-forward layers, query–key–value (QKV) projections, and normalization layers as in the VLM.
Cross-modal interaction is enabled through a shared visual key–value (KV) representation, which allows the occupancy tokens to query visual features. As illustrated in Fig.~\ref{flowchart}(a), the occupancy tokens (right) can access visual features (left) from the vision–language model via cross-attention. We can formally describe the attention operations as follows:
\begin{equation}
h_O^{occ}=O(softmax(\frac{h_Q^{occ}[h_K^{img}]^T}{\sqrt{d}})[ h_V^{img}])
\end{equation}

\begin{equation}
h_O^{img}=O(softmax(\frac{h_Q^{img}[ h_K^{img}]^T}{\sqrt{d}})[h_V^{img}])
\end{equation}

where $h_O^{img}$ denotes the image features output by the left-side of VLM, while $h_O^{occ}$ denotes the occupancy features generated by the right-side of model. Here, $h_Q, h_K$ and $h_V$ are the query, key, and value representations, and $O$ is unified output projections.  
Empirically, for the text reasoning process, we observe that whether text tokens have access to occupancy features does not result in a significant difference in quality after model convergence. 
This suggests that text can be predicted solely from visual features, indicating that during language inference, additional occupancy computation is unnecessary, thereby improving efficiency.
Finally, We insert lightweight adapters at the residual connections to fintune the VLM and preserve the original vision–language modeling capabilities.

\textbf{Latent Occupancy Prediction.} In autonomous driving scenarios, approximately 90\% of the 3D space is empty \citep{wei2024occllama}, resulting in highly sparse occupancy signals. Moreover, the raw occupancy grid is memory-intensive, typically represented as $x \in R^{H\times W\times D}$ with $ (H,W,D)=(200,200,16)$ \citep{NEURIPS2023_cabfaeec}, making direct prediction inefficient. 
We follow \citet{zheng2024occworld}, mapping the target occupancy to a compact latent space $y\in R^{\frac{H}{r} \times \frac{W}{r} \times F}$, where r is downsampling rate and F is the feature dimension of latents. As illustrated in Fig.~\ref{flowchart}, 
the left-side occupancy model outputs hidden states $h_O^{occ}$, which are projected into $z \in R^{\frac{H}{r} \times \frac{W}{r} \times F}$ via a linear projector. These features are then fed into the VQ-VAE decoder which is initialized with pretrained weights from \citet{zheng2024occworld}. Finally, a classification head converts the decoded features into the 3D occupancy predictions.

\subsection{Motion Planning}
\begin{figure}[t]
  \centering
   \includegraphics[width=1.0\linewidth]{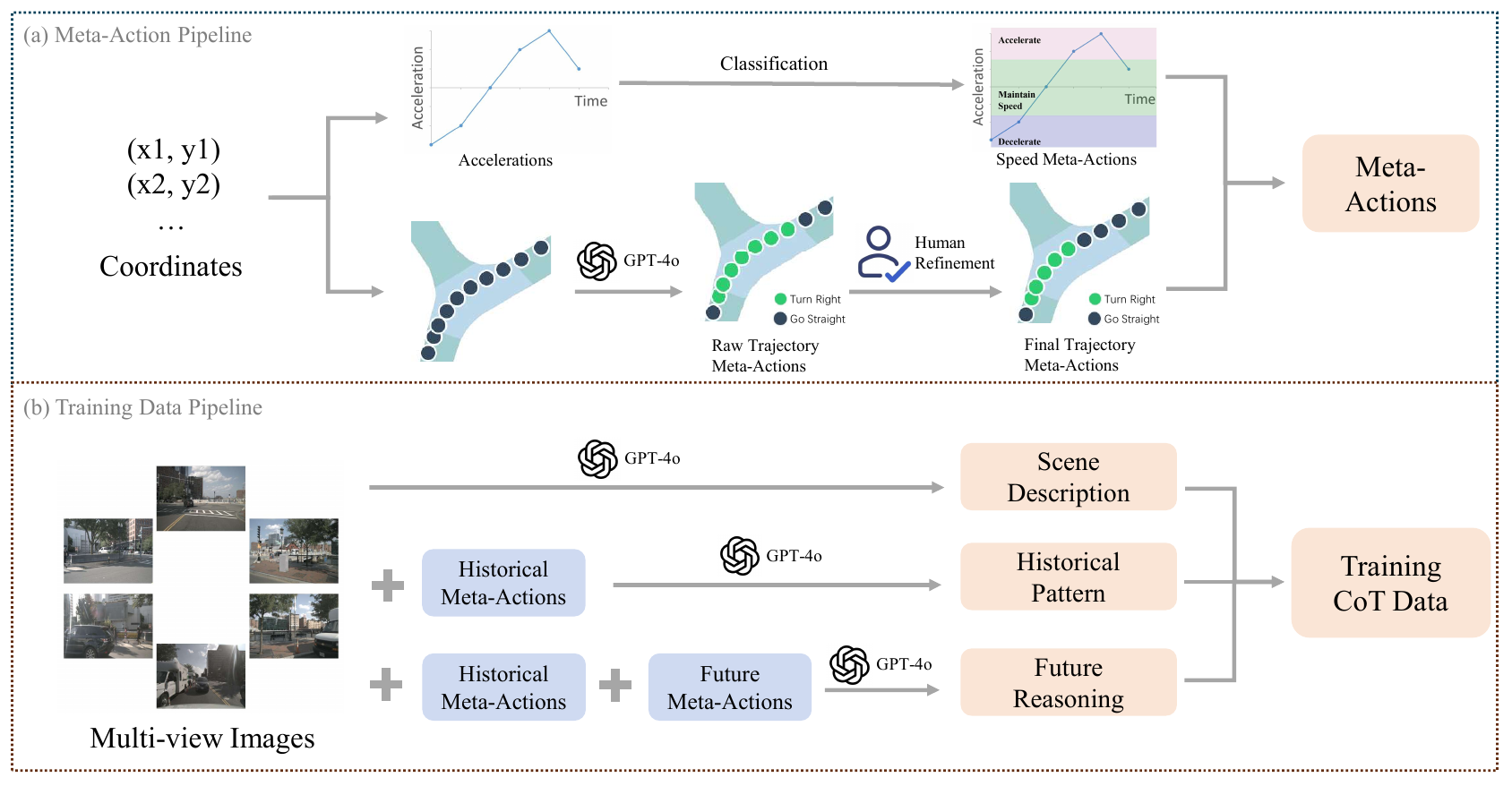}
   \caption{Overview of the meta action and CoT data generation pipeline.
(a) Meta Action Pipeline: Vehicle trajectory coordinates are processed to compute accelerations for velocity action classification, and matched to HD map lanes for trajectory action classification via GPT-4o, followed by human refinement. The two components are combined to produce final meta actions.
(b) Training Data Pipeline: Multi-view images and related meta actions are provided to GPT-4o to generate scene descriptions, infer historical motion patterns, and perform future reasoning, forming CoT training data.}
   \label{datapipeline}
   \vspace{-7mm}
\end{figure}

\textbf{Task Decomposition.} 
Large Language Models (LLMs) and Vision-Language Models (VLMs) excel at reasoning over semantic cues, but exhibit limited sensitivity to precise numerical values
.
Directly predicting future vehicle coordinates from raw trajectories therefore underutilizes their strengths. Following \citet{tian2024drivevlm}, we decompose motion planning into two stages: (1) predicting a high-level \emph{meta action} in natural language form, and (2) generating precise future coordinates using a lightweight model conditioned on the predicted meta actions.

\textbf{Meta Action Prediction.}
We define a \emph{meta action} as a compact, interpretable representation of the vehicle’s short-term driving intent, consisting of two orthogonal components: 
(1) \emph{velocity action}, categorized into \textit{Maintain speed}, \textit{Accelerate}, and \textit{Decelerate}; 
and (2) \emph{directional action}, categorized into \textit{Go Straight}, \textit{Turn Left}, \textit{Turn Right}, \textit{Change Lane Left}, \textit{Change Lane Right}, and \textit{Stop}. This formulation allows the model to reason in a discrete, language-friendly space while retaining key motion semantics.

To better utilize the reasoning capabilities of large language models, we follow \citet{hwang2024emma} and construct chain-of-thought (CoT) supervision for meta action prediction. The input to the VLM consists of six images captured from multiple perspectives, along with the past meta actions of the ego vehicle. The model first generates a natural language description of the scene, then infers the driver’s intent based on historical meta actions, and finally outputs the predicted future meta action. This multi-step reasoning encourages the model to explicitly connect scene understanding with motion intent prediction.

We develop a fully automated data construction pipeline to generate both meta action labels and their corresponding CoT annotations. For the velocity component, labels are directly obtained via threshold-based classification on acceleration. For the directional component, future trajectories are projected onto a lane-level HD map and classified by GPT-4o \citep{openai2024gpt4technicalreport} into one of the five directional categories. For the CoT annotations, GPT-4o is prompted to produce scene descriptions based on the image inputs, and then, given the ground truth meta action, to complete the reasoning steps leading to the correct label.

To ensure annotation quality, all generated meta actions on nuScenes are manually inspected, and about 20\% percent of the data has been further refined to achieve better consistency. Since the BEV perspective enables simultaneous inspection of all trajectory coordinates in a scene, minimal manual annotation effort is required. Fig.~\ref{datapipeline} demonstrates our meta action and training data pipeline.

\textbf{Planning Head.}
Given the predicted meta action, the planning head translates this high-level intent into concrete future coordinates. We adopt a simple MLP architecture inspired by \citep{li2024ego}, taking as input the meta action embedding, the previous timestep velocity, and visual tokens from the VLM. The model predicts the vehicle’s position for the next 3 seconds. Notably, no high-level navigation commands are provided, ensuring that all planning decisions emerge solely from the model’s scene understanding.

\subsection{Training Stage}
\begin{figure}[t]
  \centering
   \includegraphics[width=1.0\linewidth]{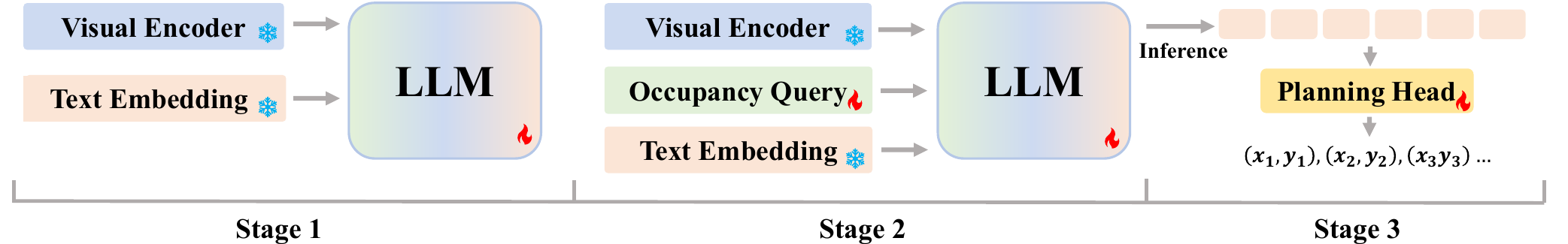}
   \caption{Overview of the training pipeline. Stage 1: Pretraining the VLM on autonomous driving scenarios using visual and text inputs. Stage 2: Occupancy-language joint training to enhance 3D scene understanding. Stage 3: Planning head training where the planning head predicts future coordinates from VLM-generated meta actions. }
   \label{stage}
   \vspace{-4mm}
\end{figure}

\textbf{Pretraining in Autonomous Driving Scenarios.}
As shown in Fig~\ref{stage}, we we adopt a VLM fine-tuning strategy along with its corresponding loss functions using the dataset sampled from OmniDrive\citep{wang2025omnidrive}. 
This phase helps the model transfer from general domains to autonomous driving scenarios, such as focusing on specific types of objects (e.g., cars, pedestrians, roads, etc.) or predicting future motion. Additionally, this training approach prepares the model to perform long-text reasoning and engage in dialogue, making it more effective in handling complex language understanding tasks.


\textbf{Occupancy-Language Joint Training.}
We focus on improving the 3D understanding capability of the VLM by aligning the Occupancy-vision modality during training. The full Occupancy-image-language dataset is used to supervise the model training, with the former eliciting 3D information representation from visual features, while the latter ensures consistency in 3D scene descriptions. To leverage the deep features of the model, we apply adapters \citep{pfeiffer2020AdapterHub, poth-etal-2023-adapters} to fine-tune the transformer blocks.
We combine the standard autoregressive language modeling loss of the LLM,  $\mathcal{L}_{ce}^{text}$ with a non-autoregressive 3D perception loss, $\mathcal{L}_{ce}^{occ}$, which calculate the cross-entropy between predicted occupancy logits and ground-truth occupancy labels. We observe that directly aligning the latent space features is suboptimal due to the inherent biases introduced by VQ-VAE encoding. Therefore, we choozaizuose to directly supervise the final 3D occupancy categories. Following  \citep{shi2025lmfusionadaptingpretrainedlanguage}, we adopt separate learning rates for different modules to further enhance training stability: the VQ-VAE decoder is assigned a learning rate of zero (rather than being fully frozen) to maintain gradient flow, while all other components share a common learning rate.

\begin{equation}
\mathcal{L}=\mathcal{L}_{AR}^{text}+\lambda\mathcal{L}_{NAR}^{occ}
\end{equation}
where $\lambda $ is a factor that controls the degree of focus on occupancy.

\textbf{Planning Head Training.}
To address the trajectory planning task, the planning head takes as input the meta actions predicted by the VLM, along with current velocity, visual tokens from the output of vlm and ego trajectories, and outputs the coordinates of the future trajectory. Specifically, the meta actions predicted by the trained VLM are fed into the planning head, whose outputs are supervised using a mean squared error (MSE) loss computed against the ground-truth trajectory coordinates.

\section{Experiment}
\subsection{Experiment Settings}

\textbf{Dataset} NuScenes is a widely used dataset in autonomous driving, consisting of 700 training scenes and 150 validation scenes. Based on the sensor information (such as images and radar) in NuScenes, Occ3D is developed as a large-scale dataset representing 3D occupancy. Furthermore, in recent years, with the advancement of large autonomous driving models, many Visual Question Answering (VQA) datasets have been built on NuScenes. We specifically evaluate the model's capabilities in 3D localization, object querying, and relational comparison using NuScenes-QA \citep{qian2023nuscenes}. Additionally, we collect a large-scale image-occupancy-text dataset to align multiple modalities and train the model to predict future meta-actions. This multimodal alignment and future prediction task aim to enhance the model's understanding of 3D scenes and its ability to reason about and act within dynamic autonomous driving scenarios.


\textbf{Implementation Details}
For all experiments, we adopt the  Paligemma2-3B-224px \citep{beyer2024paligemma, steiner2024paligemma2familyversatile} as the vision-language model backbone , while the scene VQVAE is initialized following the settings in OccWorld \citep{zheng2024occworld}. We train all models using the AdamW \citep{loshchilov2019decoupledweightdecayregularization} optimizer, and conduct experiments on 8× NVIDIA A800 GPUs. 


\subsection{Results and Analysis}

\textbf{Motion Planning}
As shown in Table \ref{t1}, we compare the motion planning capabilities of OccVLA with several strong baselines that utilize various inputs and supervisions. We observe that the current state-of-the-art method, EMMA\citep{hwang2024emma}, relies on supervision annotations (3D/BEV coordinates \& 3D bounding box), which limits its scalability to large-scale datasets. OmniDrive\citep{wang2025omnidrive}, on the other hand, depends on inputs from both camera and lidar. In contrast, OccVLA requires only camera input and uses occupancy, which can be annotated at scale, as supervision. We achieve state-of-the-art performance in terms of average L2 distance and competitive results in trajectory planning within 3 seconds.

\begin{table*}[t]
\center
\caption{End-to-end motion planning experiments on nuScenes \cite{caesar2020nuscenes} with different input and supervision. L denotes LiDAR input and C denotes camera input.}
\scalebox{1.0}{
\begin{tabular}{lllrrrr}
\toprule
\multirow{2}{*}{\textbf{Method}} & \multirow{2}{*}{\textbf{Input}} & \multirow{2}{*}{\textbf{Supervision}} & \multicolumn{4}{c}{\textbf{L2(m)$\downarrow$}}  \\
\cmidrule(lr){4-7} 
 &  &  & \textbf{1s} & \textbf{2s} & \textbf{3s} & \textbf{Avg.} \\
\midrule
NMP  & L & Box \& Motion & 0.53 & 1.25 & 2.67 & 1.48 \\
FF & L & Freespace & 0.55 & 1.20 & 2.54 & 1.43 \\
\midrule
ST-P3 & C & Map \& Box \& Depth & 1.33 & 2.11 & 2.90 & 2.11 \\
UniAD & C & Map \& Box \& Motion \& Track \& Occ & 0.48 & 0.96 & 1.65 & 1.03 \\
VAD & C & Map \& Box \& Motion & 0.54 & 1.15 & 1.98 & 1.22 \\
DriveVLM-Dual & C & Map \& Box \& Motion & 0.15 & 0.29 & 0.48 & 0.31  \\
EMMA & C & None & \textbf{0.14} & 0.29 & 0.54 & 0.32  \\

OmniDrive & C \& L & None & \textbf{0.14} & 0.29 & 0.55 & 0.33  \\
\midrule
Ours & C & Occ & 0.18 & \textbf{0.26} & \textbf{0.40} & \textbf{0.28}  \\
\bottomrule
\end{tabular}
\label{t1}
}
\vspace{-4mm}
\end{table*}

In Table \ref{t2},methods like Occ-LLM, which use occupancy as input to the LLM, encode strong 3D priors and achieve superior performance across multiple metrics. These methods use camera input and obtain Occupancy through an occupancy prediction network before feeding it into the LLM. Our method directly takes camera input and integrates the Occupancy prediction process into the LLM, achieving state-of-the-art results. Excitingly, OccVLA achieves competitive performance using only camera input compared to methods that use ground-truth Occupancy as input, further highlighting the advantage of using occupancy as an LLM output. Additionally, we achieve better performance than OccLLaMA (7B) \cite{wei2024occllama, touvron2023llamaopenefficientfoundation} with only a 3B model, demonstrating greater potential for practical applications.

\begin{table*}[ht]
\caption{End-to-end motion planning experiments on nuScenes \cite{caesar2020nuscenes} compared with models like OccNet \cite{liu2024fully}, OccWorld \cite{zheng2024occworld}, and others that use occupancy as LLM input.}
\vspace{2mm}
\centering
\scalebox{1.0}{
\begin{tabular}{lllrrrr}
\toprule
\multirow{2}{*}{\textbf{Method}} & \multirow{2}{*}{\textbf{Input}} & \multirow{2}{*}{\textbf{Supervision}} & \multicolumn{4}{c}{\textbf{L2(m)$\downarrow$}}  \\
\cmidrule(lr){4-7} 
 &  &  & \textbf{1s} & \textbf{2s} & \textbf{3s} & \textbf{Avg.}  \\
\midrule
OccNet & Occ & Map \& Box & 1.29 & 2.31 & 2.98 & 2.25  \\
OccWorld-O & Occ & None & 0.43 & 1.08 & 1.99 & 1.17  \\
OccLLAMA-O & Occ & None & 0.37 & 1.02 & 2.03 & 1.14  \\
Occ-LLM    & Occ & None & \textbf{0.12} & \textbf{0.24} & \underline{0.49} & \textbf{0.28}  \\
\midrule
OccWorld-F & C & Occ & 0.45 & 1.33 & 2.25 & 1.34  \\
OccLLama-F & C & Occ & 0.38 & 1.07 & 2.15 & 1.20  \\
Occ-LLM & C & Occ & 0.21 & 0.40 & 0.67 & 0.43  \\

\midrule
Ours & C & Occ & \underline{0.18} & \underline{0.26} & \textbf{0.40} & \textbf{0.28}  \\

\bottomrule
\end{tabular}
}
\label{t2}
\vspace{-4mm}
\end{table*}

\begin{table}[t]
\caption{Quantitative results on Nuscenes-QA\citep{qian2023nuscenes} compared with models that  using different input like LLAVA \citep{liu2023llava}, LiDAR-LLM\citep{yang2023lidarllmexploringpotentiallarge}, OccLLaMA\citep{wei2024occllama} and OpenDriveVLA\citep{zhou2025opendrivevlaendtoendautonomousdriving}.}
\centering
\vspace{2mm}
\scalebox{0.65}{
\begin{tabular}{lll ccc ccc ccc ccc ccc c}
\toprule
\multirow{2}{*}{\textbf{Model}} & \multirow{2}{*}{\textbf{Size}} & \multirow{2}{*}{\textbf{Input}} & \multicolumn{3}{c}{exist(\%)$\uparrow$} & \multicolumn{3}{c}{count(\%)$\uparrow$} & \multicolumn{3}{c}{object(\%)$\uparrow$} & \multicolumn{3}{c}{status(\%)$\uparrow$} & \multicolumn{3}{c}{comparison(\%)$\uparrow$} & \multirow{2}{*}{acc(\%)$\uparrow$} \\
\cmidrule(lr){4-6} \cmidrule(lr){7-9} \cmidrule(lr){10-12} \cmidrule(lr){13-15} \cmidrule(lr){16-18}
& & & h0         & h1         & all       & h0         & h1         & all       & h0         & h1         & all        & h0         & h1         & all        & h0           & h1          & all         &                               \\
\midrule
LLaVA 
& 7B & C &
74.8     &   72.9     &   73.8    &   14.9     &  14.3   &  14.6 &    57.7    &   34.5   &   37.9   &   48.6    &   44.5    &   45.9    &    65.9     &    52.1    &    53.3   &   47.4 
\\
LiDAR-LLM
& 7B & L & 
79.1       & 70.6       & 74.5      & 15.3       & 14.7       & 15.0      & 59.6       & 34.1       & 37.8       & 53.4      & 42.0       & 45.9       & 67.0         & 57.0        & 57.8        & 48.6
\\
OccLLaMA3.1 
& 8B & Occ & 
82.9       & 79.2       & 80.9      & 19.2       & 19.2     & 19.2      & 64.8       & 43.1       & 46.3       & 51.0       & 46.1       & 47.8       & 76.5         & 65.6       & 66.6    & 54.5
\\
OpenDriveVLA 
& 7B & C & 
-       & -       & \underline{84.2}      & -       & -     & \textbf{22.7}      & -       & -       & \underline{49.6}       & -       & -       & \underline{54.5}       & -         & -       & \textbf{68.8}    & \underline{58.2}
\\
\midrule
Ours & 3B & C &
\textbf{87.4} & \textbf{81.7} & \textbf{84.3} & \textbf{22.6} & \textbf{21.2} & \underline{21.9} & \textbf{73.6} & \textbf{51.2} & \textbf{54.5} & \textbf{62.6} & \textbf{57.9} & \textbf{59.5} & \textbf{79.2} & \textbf{66.0} & \underline{67.2} & \textbf{59.5}
\\
\bottomrule
\end{tabular}
}
\vspace{-4mm}
\label{t3}
\end{table}

\textbf{Visual Question Answering}
To further evaluate the 3D understanding capability of our model, we test it on the challenging NuScenes-QA \citep{qian2023nuscenes} benchmark. The NuScenes-QA dataset is specifically designed for autonomous driving scenarios, providing 460,000 question-answer pairs. The questions cover diverse types including existence, counting, object and status queries, and comparisons, designed to test a model's reasoning in intricate street views.

Table \ref{t3} shows the overall accuracy on NuScenes-QA. By incorporating occupancy supervision, our 3B-parameter, image-only VLM successfully outperforms larger models that rely on 3D inputs from LiDAR or explicit ground-truch occupancy data. This result highlights the superiority of our approach in fostering a deeper and more efficient 3D understanding from visual-only inputs in autonomous driving.

\begin{figure}[t]
  \centering
   \includegraphics[width=1.0\linewidth]{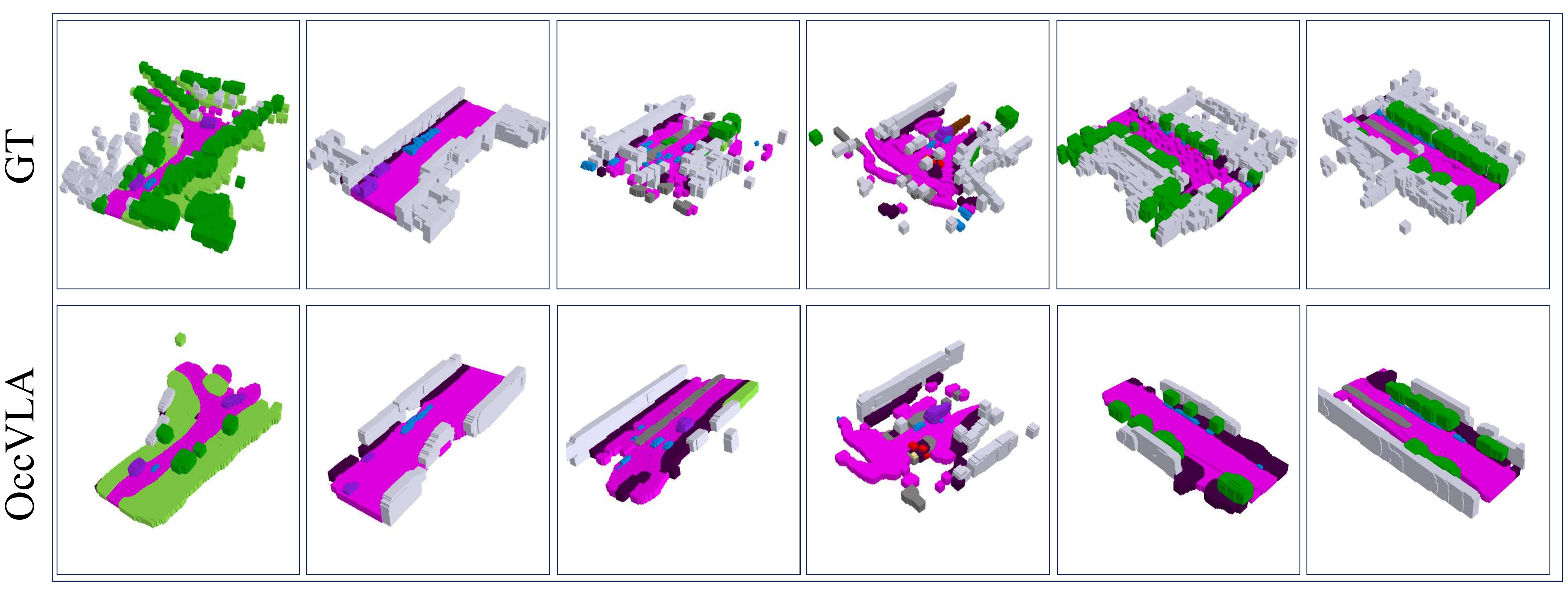}
   \caption{The 3D occupancy forecasting results of our OccVLA,  which demonstrate accurate estimation for critical objects (e.g., vehicles, roads, etc.).}
   \label{occ}
   \vspace{-4mm}
\end{figure}

\textbf{Occupancy Prediction}
The goal of this task is to predict real-time 3D occupancy using multi-view images captured by cameras. Although we employ an LLM-based architecture that is not specifically designed for occupancy prediction, our model demonstrates competitive performance, outperforming baseline methods.
Specifically, the model processes only the current time-step input without leveraging features from past states and directly outputs the 3D occupancy for the current moment, achieving about 10\% in the mIoU metric. As illustrated in the Fig.~\ref{occ}, the absence of multi-timestamp image inputs predictably limits the model's ability to handle occluded regions (e.g., buildings hidden behind trees). Nevertheless, the model excels at predicting key elements in autonomous driving scenarios, such as lanes, vehicles, pedestrians, and finer details of objects in proximity to the vehicle.

Therefore, the model exhibits a strong object-level understanding of 3D scenes in the context of autonomous driving. Despite the lack of temporal information, it effectively leverages multi-view images from the current time step to produce high-quality 3D occupancy predictions. This highlights the potential of LLM-based architectures in such tasks, even though they are not originally designed for this purpose.

\subsection{Ablation Study}


\textbf{Occupancy Supervision.} We compare the impact of occupancy prediction process on the performance of both motion planning and VQA tasks. As shown in the table, the absence of occupancy supervision means that the model relies solely on its understanding of 2D images to plan future actions. In contrast, incorporating occupancy supervision provides the model with additional 3D information, which allows it to go beyond sparse textual supervision and enhance its 3D understanding through the process of 3D occupancy prediction.
This improvement can be attributed to the occupancy supervision, which strengthens the 3D priors within the visual features learned by the LLM. Consequently, this enhancement leads to approximately a 1.5\% improvement in meta-action prediction performance.

\begin{table*}[t]
\caption{Ablation study of the occupancy supervision. The \ding{55} indicates that the model corresponds to the original VLM without occupancy integration, whereas the \ding{51} denotes that the model is trained through joint occupancy–vision–language learning.}
\centering
\vspace{2mm}
\scalebox{1.0}{
\begin{tabular}{lccccc}
\toprule
\textbf{Method} & \textbf{Occupancy}  &\textbf{speed (\%)} & \textbf{trajectory (\%)}  & \textbf{Avg. (\%)} & \textbf{Overall. (\%)}  \\
\midrule
OccVLA & \ding{55} & 53.77 & 77.24 & 65.50 & 41.48   \\
OccVLA &  \ding{51}& 54.83 & 77.95 & 66.37 & 43.08   \\

\bottomrule
\end{tabular}
}
\vspace{-4mm}
\label{a1}
\end{table*}
\vspace{-2mm}
\begin{table*}[ht]
\caption{Ablation study on Ego Trajectory. The \ding{55} symbol denotes that the model has no access to Ego Trajectory information.}
\vspace{2mm}
\centering
\scalebox{1.0}{
\begin{tabular}{lccccrr}
\toprule
\multirow{2}{*}{\textbf{Method}} & \multirow{2}{*}{\textbf{Ego Trajectory}}  & \multicolumn{4}{c}{\textbf{L2(m)$\downarrow$}}  \\
\cmidrule(lr){3-6} 
 &   & \textbf{1s} & \textbf{2s} & \textbf{3s} & \textbf{Avg.}  \\
\midrule
OccVLA  & \ding{55} & 0.28 & 0.35 & 0.80 & 0.48  \\
OccVLA  & \ding{51} & 0.18 & 0.26 & 0.40 & 0.28  \\

\bottomrule
\end{tabular}
}
\label{a3}

\end{table*}



\textbf{Ego Trajectory.} For motion planning task, previous works \citep{zhai2023rethinking, li2024ego} have raised concerns that ego trajectory might introduce excessive priors into the model, potentially leading to overfitting on the dataset. To ensure a fairer comparison, we report planning performance without past trajectory information in the table. Under the same conditions, our method demonstrates competitive performance advantages compared to state-of-the-art approaches (e.g., VAD, etc.).
Notably, our model does not rely on high-level navigation instructions; all action predictions are solely based on the model's understanding of the scene itself. This highlights the strong performance and generalization capability of OccVLA, further supporting its effectiveness in diverse scenarios.

\section{Conclusion}
In this paper, we propose OccVLA, a novel occupancy-vision-language framework for autonomous driving. OccVLA employs a parallel LLM architecture in the latent space to jointly learn occupancy and vision-language representations. This framework leverages pre-trained 2D knowledge while achieving a more critical fine-grained understanding of 3D spatial semantics. Our approach does not rely on additional 3D input information and can bypass the occupancy prediction process during inference, effectively addressing the inference delay caused by the large number of parameters in previous 3D VLM-based autonomous driving models.

\newpage
\bibliography{iclr2026_conference}
\bibliographystyle{iclr2026_conference}

\appendix

\end{document}